\documentclass{itatnew}

\makeatletter
\def\blfootnote{\xdef\@thefnmark{}\@footnotetext}
\makeatother
\usepackage[numbers]{natbib}
\usepackage{tipa}
\usepackage{graphicx}
\usepackage{tikz}
\usetikzlibrary{positioning,shapes,arrows,calc}
\usepackage{textcomp}
\usepackage{makecell}
\usepackage{enumitem}
\usepackage{pgfplots}\pgfplotsset{compat=1.9}
\usepackage{tikzscale}
\usepackage{url}
\usepackage{amsmath}
\usepackage[noabbrev,capitalize]{cleveref}
\usepackage{caption}
\usepackage[rawfloats=true]{floatrow}

\begin{document}

\title{Coarse-To-Fine And Cross-Lingual ASR Transfer}

\author{Peter Pol\'ak, Ond\v{r}ej Bojar}

\institute{  Charles University \\
Faculty of Mathematics and Physics \\
Institute of Formal and Applied Linguistics \\
\email{<polak,bojar>@ufal.mff.cuni.cz}}

\maketitle              
\blfootnote{Copyright \copyright 2021 for this paper by its authors. Use permitted under Creative Commons License Attribution 4.0 International (CC BY 4.0).}

\begin{abstract}
End-to-end neural automatic speech recognition systems achieved recently  state-of-the-art results but they require large datasets and extensive computing resources.
Transfer learning has been proposed to overcome these difficulties even across languages, e.g., German ASR trained from an English model.
We experiment with much less related languages, reusing an English model for
Czech ASR. To simplify the transfer, we propose to use an intermediate alphabet,
Czech without accents, and document that it is a highly effective strategy. The technique
is also useful on Czech data alone, in the style of coarse-to-fine training.
We achieve substantial reductions in training time as well as word error rate (WER).
\end{abstract}

\section{Introduction}

Contemporary end-to-end, deep-learning automatic speech recognition systems achie\-ved state-of-the-art results on many public speech corpora, see e.g. \citet{chiu2018state,park2019specaugment,han2019state}.

To outperform traditional hybrid models, deep-learning ASR systems must be trained on vast amounts of training data in the order of a thousand hours. Currently, there is only a limited number of public datasets that meet these quantity criteria. The variety of covered languages is also minimal. In fact, most of these large datasets contain only English \cite{wang2015transfer}. Although new speech datasets are continually emerging, producing them is a tedious and expensive task.

Another downside of new end-to-end speech recognition systems is their requirement of an extensive computation on many GPUs, taking several days to converge, see, e.g., \citet{karita2019comparative}. 

These obstacles are often mitigated with the technique of transfer learning \cite{tan2018survey} when a trained model or a model part is reused in a more or less related task.

Furthermore, it became customary to publish checkpoints alongside with the neural network implementations and 
there emerge repositories with pre-trained neural networks such as \textit{TensorFlow Hub}\footnote{\url{https://tfhub.dev/}} or \textit{PyTorch
    Hub}.\footnote{\url{https://pytorch.org/hub/}} This allows us to use pre-trained models, but similarly, most of the published checkpoints are trained for English speech.

In our work, we propose a cross-lingual coarse-to-fine intermediate step and experiment with transfer learning \cite{tan2018survey}, i.e., the reuse of pre-trained models for other tasks. Specifically,
we reuse the available English ASR checkpoint of QuartzNet \cite{kriman2019quartznet} and train it to recognize Czech speech instead.

This paper is organized as follows. In \cref{asr:related_work}, we give an
overview of related work. In \cref{sec:data_models}, we describe the used models and data. Our proposed method is described in \cref{sec:experiments}, and the results are presented and discussed in \cref{sec:results}.
Finally, in \cref{sec:conclusion} we summarize the work.

\section{Related Work}
\label{asr:related_work}

Transfer learning \cite{tan2018survey} is an established method in machine learning because many tasks do not have enough training data available, or they are too computationally demanding. In transfer learning, the model of interest is trained with the help of a more or less related ``parent'' task, reusing its data, fully or partially trained model, or its parts.

Transfer learning is gradually becoming popular in various areas of NLP. 
For example, transferring some of the parameters from parent models of
high-resource languages to low-resource ones seem very helpful in machine
translation \cite{zoph-etal-2016-transfer} even regardless the relatedness of
the languages~\cite{kocmi-bojar-2018-trivial}.

Transfer learning in end-to-end ASR is studied by
\citet{kunze-etal-2017-transfer}. They show that (partial) cross-lingual model
adaptation is sufficient for obtaining good results. Their method exploits the
layered structure of the network. In essence, they take an English model and freeze weights in the upper part of the network (closer to the input). Then they adapt the lower part for German speech recognition yielding very good results while reducing training time and the amount of needed transcribed German speech data.

More recent study on cross-lingual ASR transfer is \citet{huang2020cross}. They start with an English QuartzNet \cite{kriman2019quartznet} model trained on more than three thousand hours of transcribed speech. Further, they fine-tune the model on German, Spanish and Russian. Because the languages have a different alphabet, the authors randomly initialize a new shallow decoder. Compared with the baseline (trained from scratch), the authors report a substantial reduction in terms of WER ranging from 20~\

Other works concerning end-to-end ASR are \citet{TONG201839} and \citet{kim}. The former proposes unified IPA-based phoneme vocabulary while the latter suggests a universal character set. The first demonstrates that the model with such an alphabet is robust to multilingual setup and transfer to other languages is possible. The latter proposes language-specific gating enabling language switching that can increase the network's power.

Multilingual transfer learning in ASR is studied by \citet{cho2018multilingual}. First, they jointly train one model (encoder and decoder) on ten languages (approximately 600 hours in total). Second, they adapt the model 
for a particular target language (4 languages, not included in the previous 10, with 40 to 60 hours of training data). They show that adapting both encoder and decoder boosts the performance in terms of character error rate.

Coarse-to-fine processing \cite{raphael:coarse-to-fine:2001} has a long history in NLP. It is best known in the parsing domain, originally applied for the surface syntax \cite{charniak:etal:2006} and more recently for neural-based semantic parsing \cite{dong-lapata-2018-coarse}. The idea is to train a system on a simpler version of the task first and then gradually refine the task up to the desired complexity. With neural networks, coarse-to-fine training can lead to better internal representation, as e.g., \citet{coarse-to-fine-nmt:word-repr:2018} observe for neural machine translation.

The term coarse-to-fine is also used in the context of hyperparameter optimization, see, e.g., \citet{coarse-to-fine-hyperparam:2017} or the corresponding DataCamp class,\footnote{\url{https://campus.datacamp.com/courses/hyperparameter-tuning-in-python/informed-search?ex=1}} to cut down the space of possible hyperparameter settings quickly.

Our work is novel and differs from the above-mentioned ones in two ways: First, we
reuse existing models and checkpoints to improve the speed of training and ASR
accuracy for an unrelated language.\footnote{In contrast to the Germanic English, Czech is a Slavic language with rich morphology and relatively free word order. To the best of our knowledge, the phonetic similarity of Czech and English has not been rigorously studied, although the common belief is that Czech is more phonetically consistent.} Second, in the coarse-to-fine method,
we simplify (instead of unifying) the Czech character set to improve cross-lingual transfer and enhance monolingual training significantly.

\section{Data and Models Used}
\label{sec:data_models}

\subsection{Model architecture}

\begin{figure}[t]
    \centering
      \includegraphics[width=1\linewidth]{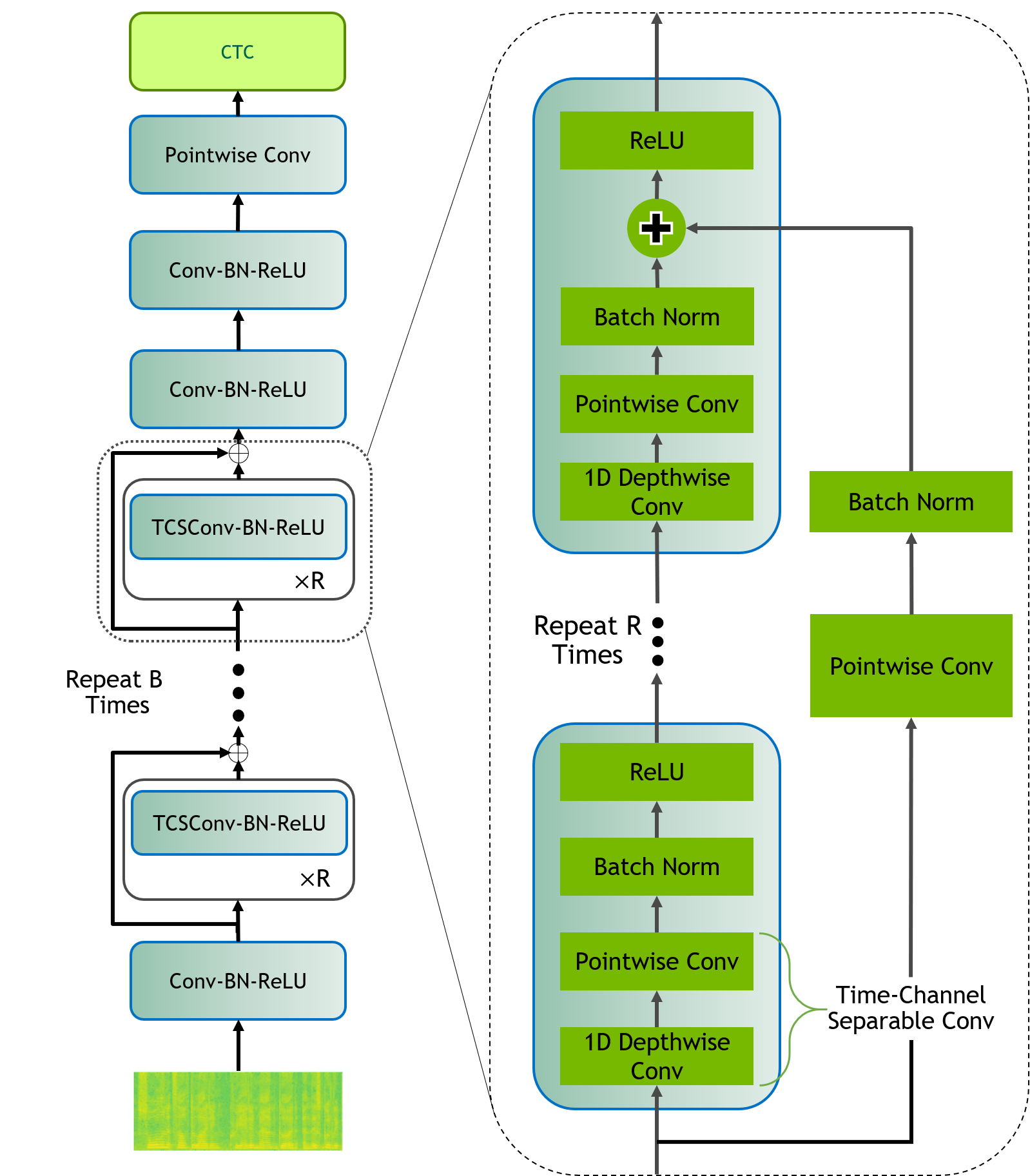}
    \caption{QuartzNet BxR architecture. Taken from \citet{kriman2019quartznet}. First layer from the input is $C_1$ layer, the last three layers are $C_2$, $C_3$ and $C_4$. From input up to the $C_3$ layer is encoder, $C_4$ layer is decoder.}
    \label{fig:quartz_arch}
\end{figure}

\begin{table}[t]
    \centering
    \begin{tabular}{lcccc}
        Block          & R & K  & $c_{out}$      & S \\ \hline\hline
        $C_1$             & 1 & 33 & 256    & 1  \\
        $B_1$             & 5 & 33 & 256    & 3 \\
        $B_2$             & 5 & 39 & 256    & 3 \\
        $B_3$             & 5 & 51 & 512    & 3 \\
        $B_4$             & 5 & 63 & 512    & 3 \\
        $B_5$             & 5 & 75 & 512    & 3 \\
        $C_2$             & 1 & 87 & 512    & 1 \\
        $C_3$             & 1 & 1  & 1024   & 1 \\
        \hline
        $C_4$             & 1 & 1  & $||$labels$||$ & 1 
    \end{tabular}
    \caption{QuartzNet 15x5 --- 5 block types ($B_1$..$B_5$) x 3 block repeats (column $S$) = 15 x 5 (column $R$)  summarized. Each block consists of $R$ $K$-sized modules. Each block is repeated $S$ times. Horizontal line marks the encoder and decoder parts of the networks.}
    \label{tab:quartz_sum}
\end{table}

We use the QuartzNet \cite{kriman2019quartznet} neural network. It is an end-to-end, convolutional neural network trained with CTC (Connectionist Temporal Classification) \cite{graves2006connectionist}, based on a larger network Jasper \cite{li2019jasper}. While performing only slightly worse then the larger Jasper, it has only a fraction of parameters (18.9 million versus 333 million).

The model input is 64 MFCC (mel-frequency cepstrum coefficient) features computed from 20 ms windows with an overlap of 10 ms. For a given time step, the model outputs probability over the given alphabet. The model starts with a 1D time-channel separable convolutional layer $C_1$ with a stride of 2. This layer is then followed by $B$ blocks. Each $i$-th block consists of the same modules repeated $R_i$ times. Each block is repeated $S_i$ times. The basic module has four layers: $K$-sized depthwise convolution layer with $c_{out}$ channels, a~pointwise convolution, a normalization layer and activation (ReLU). Next follows a time-channel separable convolution $C_2$ and two 1D convolutional layers $C_3$ and $C_4$ (with dilatation of 2). The part of the network from $C_1$ to $C_3$ is encoder and layer $C_4$ is decoder.

In our work, we use QuartzNet 15x5. This model has $5$ blocks, each repeated $S_i = 3$ times and each module repeated $R_i = 5$ times. The model schema is in \cref{fig:quartz_arch} and summarized in \cref{tab:quartz_sum}.

\subsection{Training}
We work with the original implementation in the NeMo toolkit \cite{nemo2019}.
The training is performed on 10 NVIDIA GeForce GTX 1080 Ti GPUs with 11 GB VRAM. We use the \texttt{O1} optimization setting, which primarily means mixed-precision training (weights are stored in single precision, gradient updates are computed in double precision). Batch size is 32 per GPU ($10 \times 32 = 320$ global batch size). We use warm-up of 1000 steps. The learning rate for training is $0.01$ and $0.001$ for the fine-tuning. The optimizer is NovoGrad \cite{ginsburg2019stochastic} with weight decay 0.001 and $\beta_1 = 0.95$ and $\beta_2 = 0.5$. Additionally, we use Cutout \cite{devries2017improved} with 5 masks, maximum time cut 120 and maximum frequency cut 50.

\subsection{Pre-Trained English ASR}
As the parent English model, we use the checkpoint available at the \textit{NVIDIA GPU Cloud}.\footnote{\url{https://ngc.nvidia.com/catalog/models/nvidia:quartznet15x5}} It is trained on LibriSpeech \cite{panayotov2015librispeech} and Mozilla CommonVoice\footnote{\url{https://commonvoice.mozilla.org/en/datasets}} for 100 epochs on 8 NVIDIA V100 GPUs. The model achieves 4.19\,\

\subsection{Czech Speech Data}
In our experiments, we use Large Corpus of Czech Parliament Plenary Hearings \cite{kratochvil2020large}. At the time of writing, it is the most extensive available speech corpus for the Czech language, consisting of approximately 400 hours.

The corpus includes two held out sets: the development set extracted from the training data and reflecting the distribution of speakers and topics, and the test set which comes from a different period of hearings. We choose the latter for our experiments because we prefer the more realistic setting with a lower chance of speaker and topic overlap.

\section{Examined Configurations}
\label{sec:experiments}

\begin{figure*}[t]
    \centering
    \includegraphics[width=0.8\linewidth]{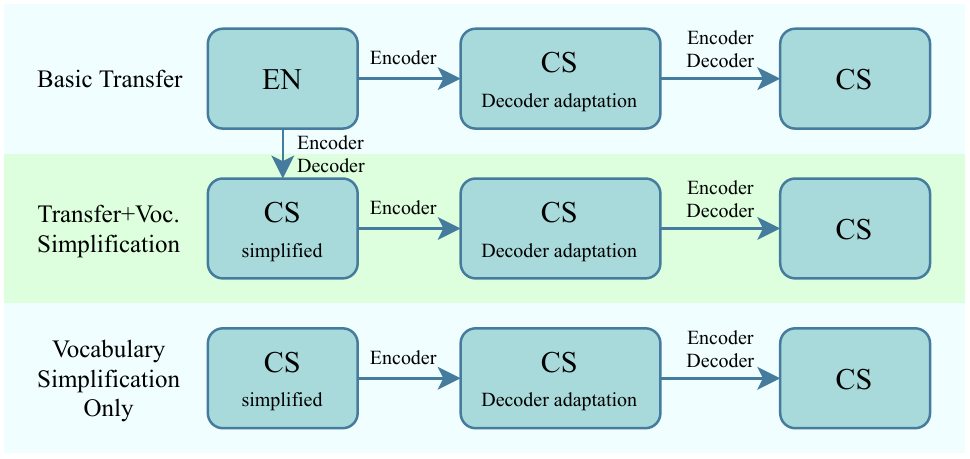}
    \caption{Examined setups of transfer learning. The labels on the arrows
    indicate which model parts are transferred, i.e., used to initialize the
    subsequent model. Parameter freezing is involved only for the encoder weights in the ``CS decoder adaptation'' phase.}
    \label{fig:transfers}
\end{figure*}

\cref{fig:transfers} presents the examined setups. In all cases, we aim at the best possible Czech ASR, disregarding the model's performance in the original English task. The baseline (not in the diagram) is to train the network from scratch on the whole Czech dataset, converting the speech signal directly to Czech graphemes, i.e., words in fully correct orthography, except punctuation and casing which are missing in both the training and evaluation data.

\subsection{Basic Transfer Learning}
\label{basic_transfer}

Our first method is very similar to \citet{kunze-etal-2017-transfer} and \citet{huang2020cross}. We use the English checkpoint with the (English) WER of 4.19\,\

The Czech language uses an extended Latin alphabet, with diacritic marks (acute,
caron, and ring) added to some letters. The Czech alphabet has 42 letters, including the digraph ``ch''. Ignoring this digraph (it is always written using the letters ``c'' and ``h''), we arrive at 41 letters. Only 26 of them are known to the initial English decoder.

To handle this difference, we use a rapid decoder adaptation (unlike \citet{huang2020cross}). For the first 1500 steps, we keep the encoder frozen and only train the decoder (randomly initialized; Glorot uniform).

Subsequently, we unfreeze the encoder and train the whole network on the Czech dataset.

\subsection{Transfer Learning with Vocabulary Simplification}

In this experiment, we try to make the adaptation easier by first keeping the original English alphabet and extending it to the full Czech alphabet only once it is trained.

To coerce Czech into the English alphabet, it is sufficient to strip diacritics (e.g. convert ``\v{c}\'arka'' to ``carka''). This simplification is quite common in Internet communication but it always conflates two sounds (\textipa{[ts]} written as ``c'' and \textipa{[tS]} written as ``\v{c}'')  or their duration (\textipa{[a:]} for ``\'a'' and \textipa{[a]} for ``a'').

In this experiment, we first initialize both encoder and decoder weights from the English checkpoint (English and simplified Czech vocabularies are identical so the decoder dimensions match), and we train the network on the simplified Czech dataset for 39 thousand steps.

The rest (adaptation and training on the full Czech alphabet)
is the same as in \cref{basic_transfer}.

Overall, this can be seen as a simple version of coarse-to-fine training where a single intermediate model is constructed with a reduced output alphabet.

\subsection{Vocabulary Simplification Only}
\label{sub_sec:simplification}

In this experiment, we first simplify target vocabulary: we use standard Latin alphabet with 26 letters plus space and apostrophe (to preserve compatibility with English). Czech transcripts are then encoded using this simplified alphabet (e.g. ``\v{c}\'arka'' as ``car\-ka'').

With transcripts encoded in this manner, we train a randomly (Glorot uniform) initialized QuartzNet network for 39 thousand steps. 

From our previous experience with vocabulary adaptation, we make a short adaptation of the model for a different alphabet. We initialize the encoder with weights obtained in the previous step and modify the target vocabulary to all Czech letters (41 plus space and apostrophe). The decoder is initialized with random weights. We freeze the encoder and train this network shortly for 1500 steps. Note that the original decoder for simplified Czech is discarded and trained from random initialization in this adaptation phase.

After this brief adaptation step, we unfreeze the encoder and train the whole network for 39 thousand steps.

\begin{figure*}[t]
\includegraphics[width=\linewidth,height=60mm]{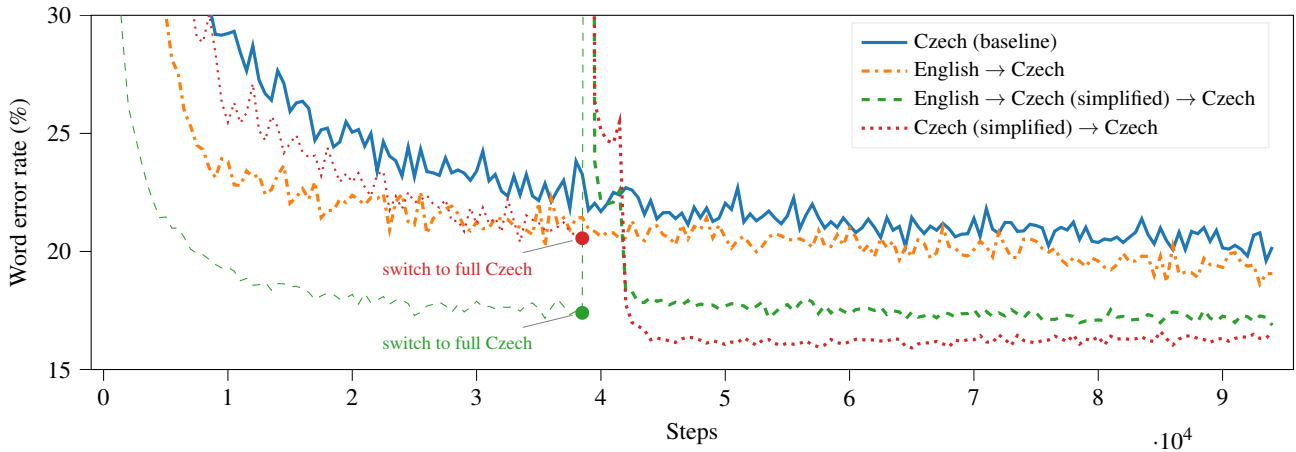}
\caption{Evaluation on the test set during training. 
In setups with transfer learning, the same colour and dashing are used across training stages.
Note that WER curves for experiments with simplified vocabulary (thin lines)
are not directly comparable with other curves until step 39,000 as the test set
is on different (simplified) vocabulary. 10,000 steps take approximately 5 hours. Better viewed in colour.}
\label{fig:training}
\end{figure*}

\begin{table}[t]
    \centering
    \small
    \begin{tabular}{lc|cc}
        
        \bf Experiment & \bf Simplified & \bf Adaptation & \bf Full \\
        \hline
        Baseline CS & - &  - &  20.19 \\
        
        EN $\rightarrow$ CS & -  & 97.35 &  19.06  \\
        
        EN $\rightarrow$ sim. CS $\rightarrow$ CS & 17.11  & 22.01 &  16.88 \\
        
        sim. CS $\rightarrow$ CS & 20.56  &  24.59 &  16.57  \\
        
    \end{tabular}
    \caption{Results in~\
    column reflects WER after the training on simplified dataset (both training and test data without accents). ``Adaptation'' column contains WER immediately after the decoder adaptation phase to the full Czech alphabet including accents.
    Finally, ``Full'' column contains performance on the test set with accents 
    after the full training.}
    \label{tab:results}
\end{table}

\section{Results and Discussion}
\label{sec:results}

\cref{tab:results} presents our final WER scores and \cref{fig:training} shows their development through the training. For simplicity, we use greedy decoding and no language model.
We take the model from the last reached training iteration. Arguably, some of the setups are not yet fully converged but the main goal of this paper is to propose solutions for situations where hardware resources are capped (as was our case, too). With unlimited training time available, the results might differ.
Little signs of overfitting are apparent for the ``Simplified CS $\rightarrow$ CS'' setup. An earlier iteration of this setup might have worked better, but we do not have another test set with unseen speakers to validate it. The development set of the Czech speech corpus is rather small (3 hours), so we prefer not to split it.

Without an independent test set and computing capacity to reach full convergence or overfitting for all the models, our analysis has to focus on the development of model performance in time (\cref{fig:training}) rather than on the final performance of models chosen by an automatic stopping criterion on such a held-out test set.

\subsection{Transfer across Unrelated Languages}

We observe that initialization 
with an unrelated language helps to speed up training.

This is best apparent by comparing the first 39k steps of the learning curves for ``English
$\rightarrow$ Czech simplified'' and ``Czech (simplified) $\rightarrow$ Czech''
in \cref{fig:training}, thin lines. Here the target alphabet was simplified in
both cases, but the weights
initialized from English allow a much faster decrease of WER. The benefit is also clear for the full alphabet (baseline vs. ``English $\rightarrow$ Czech''), where
the baseline has a consistently higher WER.

The training of the English parent is so fast that WER for the
simplified alphabet (thin dashed green line) drops under 30\,\
steps (1 hour of training).

This can be particularly useful if the final task does not require the lowest possible WER, such as sound segmentation.

While Basic transfer (``English $\rightarrow$ Czech'') boosts the convergence
throughout the training, its final performance is only 1 to 2\,\
better than the baseline, see the plot or compare 20.19 with 19.06 in
\cref{tab:results}. The intermediate vocabulary simplification is more important
and allow a further decrease of 2.2\,\
from English.

\subsection{Transfer across Target Vocabularies}

In the course of two experiment runs, we altered the target vocabulary: the
training starts with simplified Czech, and after about 39,000 steps, we switch
to the full target vocabulary. This sudden change can be seen as spikes in
\cref{fig:training}. Note that WER curves before the peak use the simplified
Czech reference (the same test set but with stripped diacritics), so they are
not directly comparable to the rest.

The intermediate simplified vocabulary always brings a considerable improvement. In essence, the final WER is lower by 2.18 (16.88 vs. 19.06 in \cref{tab:results}) for the models transferred from English and by 3.62 (16.57 vs. 20.19) for Czech-only runs.
One possible reason for this improvement is the ``easier'' intermediate task of
simpler Czech. Note that the exact difficulty is hard to compare as the target
alphabet is smaller than with the full vocabulary, but more spelling ambiguities may arise. This intermediate
task thus seems to help the network to find a better-generalizing region in the parameter space. Another possible reason that this sudden change and reset of the last few layers allows the model to reassess and escape a local optimum in which the ``English $\rightarrow$ Czech'' setup could be trapped.

\section{Conclusion and Future Work}
\label{sec:conclusion}

We presented our experiments with transfer learning for automated speech recognition between unrelated languages.
In all our experiments, we outperformed the baseline in terms of speed of convergence and accuracy.

We gain a substantial speed-up when training Czech ASR while reusing weights
from a pre-trained English ASR model. The final word error rate improves over
the baseline only marginally in this basic transfer learning setup.

We are able to achieve a substantial improvement in WER by introducing an intermediate step in the style of coarse-to-fine training, first training the models to produce Czech without accents, and then refining the model to the full Czech.
This coarse-to-fine training is most successful within a single language: Our final model for Czech is better by over 3.5 WER absolute over the baseline, reaching WER of 16.57\

As we documented in \cref{sec:results}, transfer learning leads to a substantial reduction in training time. We achieved speed-up even in unrelated languages. We also demonstrated that the coarse-to-fine approach leads not only to training time reduction but also yields better accuracy.

We see further potential in the coarse-to-fine training. We want to explore this area more thoroughly, e.g., by introducing multiple simplification stages or testing the technique on more languages.

\section*{Acknowledgements}

The work was supported by the grant
 19-26934X (NEUREM3) of the Czech Science Foundation 
and START/SCI/089 (Babel Octopus: Robust Multi-Source Speech Translation) of the START Programme of Charles University.

\bibliographystyle{acl_natbib}

\end{document}